\documentclass[runningheads]{llncs}
\usepackage{xcolor}
\usepackage{makecell}
\usepackage{graphicx}

\usepackage{tabularx}
\begin{document}
%
\title{Detecting abnormalities in resting-state dynamics: An unsupervised learning approach}
\titlerunning{Unsupervised anomaly detection to detect outliers in rs-fMRI}  
%

\author{Meenakshi Khosla \inst{1} \and Keith Jamison \inst{2,3} \and Amy Kuceyeski \inst{2,3} \and Mert R. Sabuncu \inst{1,4}}

\institute{1. School of Electrical \& Computer Engineering, Cornell University\\
2. Radiology, Weill Cornell Medical College \\
3. Brain and Mind Research Institute, Weill Cornell Medical College \\
4.  Nancy E. \& Peter C. Meinig School of Biomedical Engineering, Cornell University}

%
%
%

%
\maketitle              
\begin{abstract}
Resting-state functional MRI (rs-fMRI) is a rich imaging modality that captures spontaneous brain activity patterns, revealing clues about the connectomic organization of the human brain.
While many rs-fMRI studies have focused on static measures of functional connectivity, there has been a recent surge in examining the temporal patterns in these data.
In this paper, we explore two strategies for capturing the normal variability in resting-state activity across a healthy population: (a) an autoencoder approach on the rs-fMRI sequence, and (b) a next frame prediction strategy. 
We show that both approaches can learn useful representations of rs-fMRI data and demonstrate their novel application for abnormality detection in the context of discriminating autism patients from healthy controls.


\end{abstract}
\section{Introduction}
Resting-state fMRI captures intrinsic neural activity, in the absence of external stimuli and task requirements. Much of the research in this direction has aimed at identifying connectivity based biomarkers, restricting the analysis to so-called ``static'' functional connectivity measures that quantify the \textit{average} degree of synchrony between brain regions. 
For e.g., machine learning based strategies have been used with static connectivity measures to parcellate the brain into functional networks, and extract individual-level predictions about cognitive state or clinical condition~\cite{khosla2018machine}.
In recent years, there has been a surge in the study of the temporal dynamics of rs-fMRI data, offering a complementary perspective on the functional connectome and how it is altered in disease, development, and aging~\cite{tian2018changes}.
However, to our knowledge, there has been a dearth of machine learning applications to dynamic rs-fMRI analysis.


Thanks to large-scale datasets, modern machine learning methods have fueled significant progress in computer vision. Compared to natural vision applications, however, medical imaging poses a unique set of challenges. Data, particularly labeled data, are often scarce in medical imaging applications. This makes data-hungry methods such as supervised CNNs possibly less useful. One potential approach to tackle the limited sample size issue is to exploit unsupervised or semi-supervised learning strategies that don't depend on large amounts of labeled training data.
In this paper, we explore the use of unsupervised end-to-end learning for capturing rs-fMRI dynamics and demonstrate that the representations our models learn can be useful for detecting abnormal patterns in data. 





\paragraph{\textbf{Related Work:}}

Machine learning methods are increasingly used to compute individual-level predictions from rs-fMRI data, e.g. about disease~\cite{khosla2018machine}. The conventional approach of supervised learning relies on labeled training data and uses hand-crafted features such as the static correlation between pairs of regions. Such features fail to capture the dynamics of resting-state activity as it relates to behavior or disease. Moreover, emerging data suggest that learning models that exploit the full-resolution 4-dimensional fMRI data can potentially reveal more discriminative resting-state biomarkers~\cite{liu2018chronnectome}. In this work, we are motivated by this observation and our goal is to move away from hand-crafted features and take full advantage of the spatio-temporal structure of rs-fMRI.

Unsupervised approaches such as clustering of static connectivity measures have been previously used for disease classification and discovery of novel disease sub-types~\cite{pmid23616377}.  Similarly, autoencoders have been used in pre-training to improve generalization capabilities of supervised learning algorithms, as in~\cite{suk2015hybrid}. 
An alternative application of unsupervised learning is outlier detection. Here, the goal is to identify data points that deviate markedly from normal samples. For example, autoencoder models have been popular for outlier detection in video~\cite{Hasan2016LearningTR}. In recent years, predictive modeling has also been shown to be a powerful framework in unsupervised feature learning of video representations~\cite{Srivastava2015UnsupervisedLO}. 
In this approach, a model is trained to predict future frames of a video sequence. These models learn useful internal representations of the data that can in turn be used for anomaly detection or downstream object recognition or classification tasks \cite{Liu2018FutureFP}. 

In the present paper, we propose a novel unsupervised approach that learns rs-fMRI representations on voxel-level time-course data captured via a convolutional RNN model, in an end-to-end learning fashion.
Models are trained to predict the next frame in an rs-fMRI sequence or to reconstruct the entire sequence. We apply our approach to the novel problem of outlier detection in rs-fMRI, and demonstrate its utility in discriminating autism patients from healthy controls.



%
%
\section{Methodology}

In this section, we describe the autoencoder and prediction models considered in the study. As we demonstrate empirically, the models learn to accurately reconstruct or predict “normal” resting-state activity in healthy subjects, but yield higher reconstruction/prediction errors in patients.
%

\subsection{Network building blocks}
\paragraph{Convolutional networks:}
CNNs have achieved unprecedented levels of performance across many vision tasks~\cite{krizhevsky2012imagenet}. 
The main ingredients of CNNs include convolutional layers that serve as feature extractors, and pooling/un-pooling layers that perform down/up-sampling in resolution.
In this paper, we employ encoder-decoder style networks since we are reconstructing/predicting structured image data, i.e., rs-fMRI frames. Encoder-decoder networks are widely deployed in image segmentation and generation tasks, as in~\cite{Ronneberger2015UNetCN}. 
The encoding part computes a cascade of increasingly high-level representations from the images, 
whereas the decoding part reconstructs pixel-level features from these representations. 

\paragraph{Convolutional-LSTM networks:}
Recurrent neural networks (RNNs), e.g., LSTMs~\cite{hochreiter1997long}, offer state-of-the-art results in many domains with sequential data, such as speech or natural language processing. 
Conv-LSTM cells, an extension of LSTM units, integrate convolutional layers with LSTM modules and allow the temporal propagation of high-level spatial features captured by convolutional layers. 
Conv-LSTM cells have shown remarkable performance in sequence forecasting problems \cite{Shi2015ConvolutionalLN}. This stems from their ability to simultaneously capture rich spatial and temporal structures in the data.

\subsection{Next frame prediction model}
\begin{figure}
\includegraphics[width=0.9\textwidth]{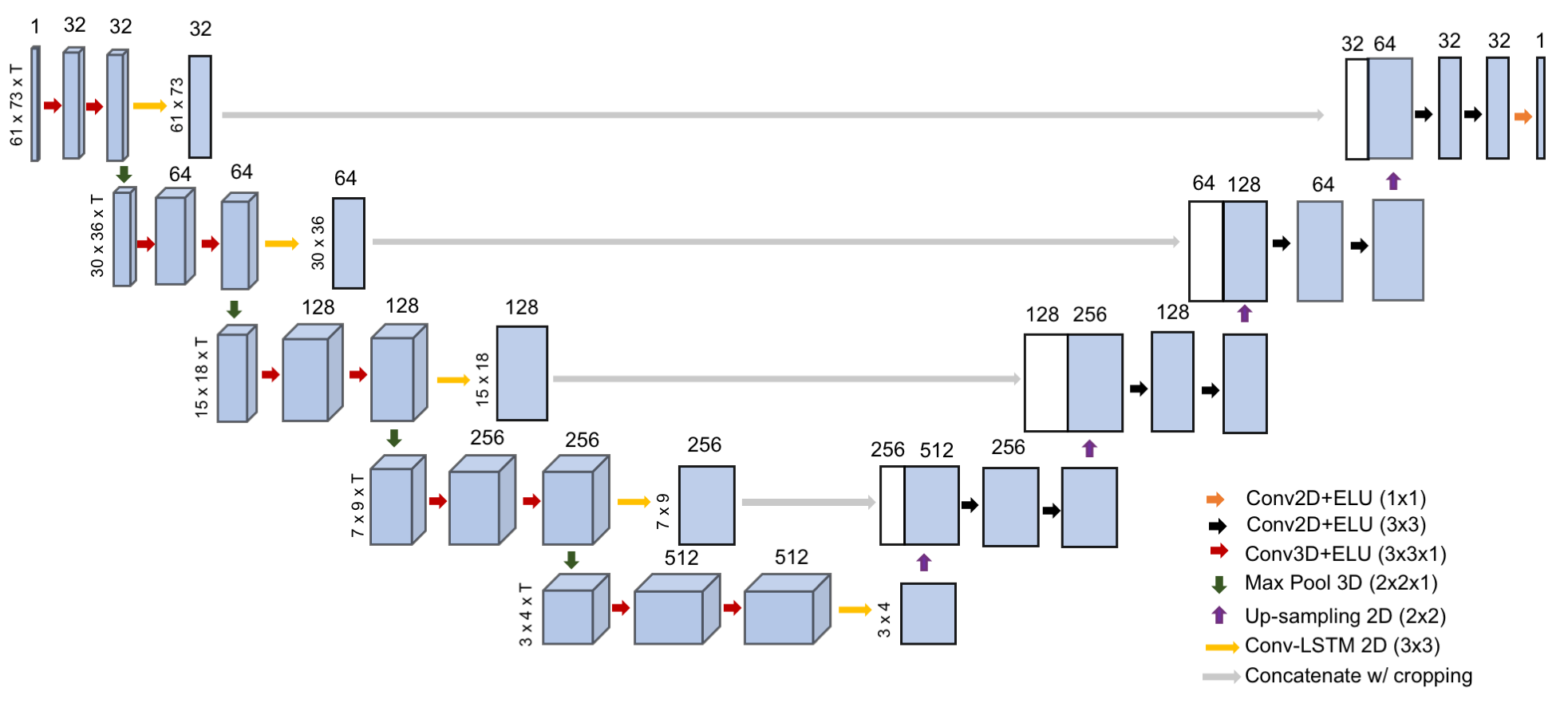}
\caption{Next frame prediction model. Each cuboid represents a 3D (2 spatial dimensions +  time) feature map with number of features indicated on top. Flat boxes represent 2D feature maps, with number of channels on top. Input is an axial fMRI slice with T sequential frames. Conv-LSTM cell returns the last output of the output sequence.} \label{fig0}
\end{figure}

Given a sequence of rs-fMRI frames, we trained a model to predict the next frame in the sequence. To improve the localization accuracy of predicted frames and capture spatio-temporal correlations at multiple resolutions, we incorporate skip connections with Conv-LSTM modules in our architecture. This U-Net style architecture~\cite{Ronneberger2015UNetCN} is shown in Figure \ref{fig0}. 
The input to the  model is a 2D rs-fMRI sequence of $T$ axial slices.
In the encoding layers, we used 3D convolutions and max pooling, where the first two dimensions are the spatial coordinates on the axial cross-section and the third dimension is time.
We compared our prediction model with several baselines, including: (a) simply using the last frame of the input sequence as a prediction of the next frame;
(b) a non-learning based extrapolation model that fits separate cubic splines at each pixel on the input sequence; and
(c) a non-recurrent 2-D U-Net model that excludes the Conv-LSTM modules from the proposed architecture and treats the temporal component of the input as T channels. 
We also considered (d) an interpolation scheme that interpolated with cubic splines between the $T$ frames of the input sequence that precede the predicted frame \textit{and} the frame \textit{after} the predicted frame. This interpolation method is different than the other methods as it is not a forecasting model, yet we found it useful to assess the performance of the other methods.

\subsection{Autoencoder model}
The autoencoder is an unsupervised learning approach that encodes the input into a lower dimensional representation, which is then decoded into a reconstruction of the input.
The model is trained to minimize a distance function between the reconstruction and input, such as the squared $L_2$ distance. 
The architecture of our reconstruction model is the same as the prediction model above, with two important differences. First, there are no skip connections, which are indicated as a ``concatenate with crop'' operation, to avoid the trivial solution of copying input to the output.
The second difference is that, in the decoder layers and the output we have $T$ frames, instead of a single frame. So in the visualization of this architecture, those would be represented with cuboids and 3D convolution/up-sampling operations. Further, we retained Conv-LSTM unit in the bottleneck to capture temporal dependencies between the frames of a rs-fMRI sequence. 

\section{Experiments}
\subsection{Data}
We conducted our experiments on data from the Autism Brain Imaging Data Exchange (ABIDE) study~\cite{di2014autism}. Because of difference in TRs and other imaging parameters across sites, we restricted our experiments to the acquisition site with the largest sample size, namely NYU. We only used data that passed quality assessments by all functional raters and retained enough time-points after motion scrubbing for band-pass filtering. We randomly selected two thirds of the healthy group (54 subjects) for training/validating the reconstruction \& imputation models. A validation split of 10\% was used during training to monitor convergence of these models. The remaining one-third group comprising 28 healthy controls was used as test data to evaluate predictions/reconstruction performance for comparison against ASD patients (N=67).   


Rs-fMRI preprocessing included slice timing correction, motion correction, global mean intensity normalization, standardization of functional data to MNI space, global signal regression, motion scrubbing (volume censoring) and band-pass filtering. We note that band-pass filtering was performed after motion scrubbing to avoid any motion contamination. Individual rs-fMRI scans were normalized between 0 to 1 by min-max scaling each-individual voxel's time series.
Finally, we applied a binary gray matter mask to all 3D volumes \cite{AAL}.

\subsection{Implementation Details}
During training, we identified non-overlapping contiguous segments of $(T+1)$ frames for each subject in the training set. For each such segment, we extracted all axial slices and trained a unified model to predict the next frame, i.e, for  a given architecture a single model was trained for all subjects and axial slices, comprising 16,560 training instances. 
Squared loss was optimized with Adam and a learning rate 1e-4 and AMSGrad. We implemented our code using Keras, with a TensorFlow back-end. The network was trained for 150 epochs with a batch size of 32. Validation curves were monitored to ensure convergence. We used same training paradigm for the non-recurrent baseline U-Net model. In our experiments, we tried different values for $T$ and observed diminishing returns beyond $T=20$ in the performance of the next frame prediction models. The overall pattern in comparing the accuracy of different models was the same. Thus, in the remainder we fix $T=20$. We note that, while not necessary, we fixed $T=20$ for the autoencoder models too, which ensured  training was done on identical datasets for these different approaches. Once the models were trained, we used them to compute predictions or reconstructions on independent data, which included both controls and ASD patients.
For each test subject, we computed  the  mean squared error (between reconstruction/prediction and ground truth frames) as a single metric. Note that we averaged over all frames and pixels in an rs-fMRI scan. We hypothesized that this metric would be different between patients and controls, demonstrating that it could be used as an outlier detector. We also analyzed the voxel-level squared errors and conducted a statistical comparison between patients and controls to reveal the anatomical distribution  of the differences. 

\section{Results}
\subsection{Next Frame prediction and reconstruction errors}
We first demonstrate that the next frame in rs-fMRI sequence can be accurately predicted. Table~\ref{tb:1} shows the performance of the different  methods we implemented. 
We list both mean squared error and the average Pearson's correlation between predicted and ground truth frames, computed within the gray matter mask on healthy test subjects.
We observe that the proposed recurrent U-Net architecture achieves the best prediction performance, even exceeding the cubic-spline based interpolator, which was given both the preceding 20 frames and the frame \textit{after} the predicted frame. 
The recurrent LSTM modules that capture the temporal dynamics also enabled a significant boost in quality, as can be noted by comparing the performance of the U-Net and proposed architecture.
Finally, the U-Net models outperformed the non-learning based methods of extrapolation, suggesting that accounting for both the spatial and temporal structure in the data yielded better results.

Table~\ref{tb:rec} shows the mean reconstruction errors of the autoencoder on healthy test subjects for various input sequence lengths at test time. We note that the performance is worse than next-frame prediction because of the absence of skip connections. Reconstruction quality degraded with fewer frames suggesting that the autoencoder is not reconstructing frames independently and is indeed exploiting the long-term temporal dependencies between frames. For outlier detection, we thus used the temporal window T=20 as it gives the best reconstruction performance and captures longer dynamics.     
\begin{table}
\begin{center}
\begin{tabular}{| c | c | c |}
\hline
Imputation models & Mean Squared Error & Pearson's Correlation \\
\hline
\textbf{Last observation copy} &  0.01969  & 0.7558 \\
\textbf{Extrapolation} &  0.01203 & 0.8938 \\
\textbf{Interpolation*} & 0.00065 & 0.9939 \\
\textbf{Non-recurrent U-Net} & 0.00026 & 0.9967 \\
\textbf{Proposed recurrent U-Net} &  0.00007 & 0.9990 \\
\hline
\end{tabular}
\caption{Next frame prediction performance on healthy test subjects for different models. *Interpolation model had access to the frame after the predicted frame.  \label{tb:1}}
\end{center}
\end{table}
\vspace{-0.5in}
\begin{table}
 \begin{center}
     \begin{tabular}{| c | c | c |}
     \hline 
  Recurrent autoencoder: sequence length &  Mean squared error & Pearson's correlation \\
\hline 
     \textbf{T=10 frames}  & 0.0625 & 0.354  \\
      \textbf{T=15 frames}  &  0.0475 & 0.503  \\
       \textbf{T=20 frames}  &  0.0437 & 0.550 \\
     \hline
     \end{tabular}
     \caption{Reconstruction performance of the proposed recurrent autoencoder on healthy test subjects for different input sequence lengths. \label{tb:rec}}
 \end{center}
 \end{table}
\vspace{-0.4in}
\subsection{Outlier Detection: Discriminating Patients and Controls} 
\begin{figure}[t]
\includegraphics[width=0.49\textwidth]{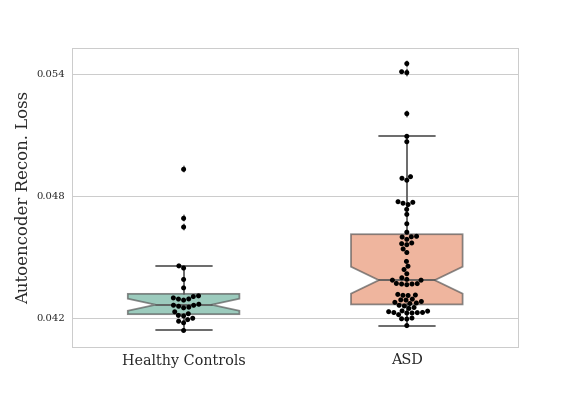}
\includegraphics[width=0.49\textwidth]{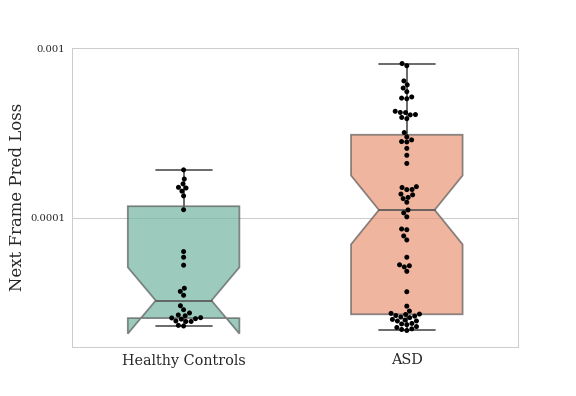}
\caption{Whisker plots showing reconstruction and prediction errors (mean squared error) for ASD patients and controls, with proposed recurrent models trained on T=20 consecutive frames. Points are individual subjects. The ends of the box are upper and lower quartiles, the median is marked by a horizontal line inside the box.} \label{fig1}
\end{figure}

We were interested in examining whether the next frame prediction and reconstruction models can be used to detect outlier subjects.
To test this, we computed mean squared error on all test subjects, including healthy controls and ASD patients.
Figure~\ref{fig1} shows these error values for the proposed next frame prediction and autoencoder models. 
Both models yield error values that are statistically significantly different between the two clinical groups. Further, AUC values obtained with autoencoder and imputation models, as shown in Table~\ref{tb:auc}, are on par with recent supervised ASD v/s control classification results~\cite{abraham2017deriving}. 

\vspace{-0.1in}
 \begin{table}
 \begin{center}
 \begin{tabular}{| c | c |  }
 \hline
      Model & AUC (p-value)\\
       \hline 
 Recurrent autoencoder  & 69.6 (0.00466) \\
 U-Net imputation &  62.5 (0.00293) \\
 Recurrent U-Net imputation & 65.9 (0.00151) \\
\hline 
     \end{tabular}
     \caption{Area under the ROC curve for discriminating ASD vs Controls. P-values of the unpaired t-test comparing means of the two clinical groups are shown in brackets. \label{tb:auc}}
 \end{center}
 \end{table}

 \vspace{-0.3in}


We also note that the non-recurrent U-Net benchmark achieves a weaker separation between the two clinical groups. This indicates that the conv-LSTM layers enhance diagnostic sensitivity presumably because they are more equipped to exploit spatiotemporal structure in extracting representations.
Importantly, we observed no correlation between frame-wise displacement values (a widely used metric to quantify subject motion) and the prediction/reconstruction errors- neither at the frame-level (Pearson's correlation -0.0161/0.0218, p = 0.0739/0.0251, computed on non-motion scrubbed frames only) nor at the individual level (Pearson's correlation 0.0033/0.1730, p = 0.9744/0.0936).

\begin{figure}
\begin{center}
\includegraphics[width=0.8\textwidth]{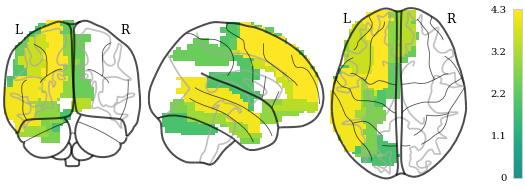}
\caption{Statistical significance of the difference in regional reconstruction error of the recurrent autoencoder between controls and ASD patients. FDR with $q=0.05$ was implemented for multiple testing correction. $-\log_{10}$ p values are shown.} \label{fig:sigmap}
\end{center}
\end{figure}
Finally, we were interested in exploring the anatomical differences in errors between the two clinical groups. We thus conducted a t-test of of the regional prediction error (averaged within the boundaries of the widely used AAL atlas \cite{AAL}) on the model with best AUC, i.e. the autoencoder. As can be seen from Fig~\ref{fig:sigmap}, significant differences were mainly constrained to the left hemisphere, particularly localizing within the language network, involving the temporal and frontal cortices, consistent with prior literature~\cite{eyler2012failure}. 
\section{Discussion}
We considered a novel unsupervised learning strategy to analyze resting-state fMRI data, where we train recurrent models to reconstruct rs-fMRI clips or to predict the next frame in the sequence. Results indicate that the proposed recurrent U-Net architecture produces very accurate predictions that yield a correlation greater than $0.99$ with ground truth frames. Furthermore, this performance is better than an interpolation approach that had access to the frame after the predicted frame. Next, we demonstrated the utility of the proposed models in detecting outliers in rs-fMRI. Our results indicate that next frame prediction error or reconstruction error can be used to discriminate patients from healthy controls, achieving a classification performance close to state-of-the-art results obtained with supervised methods. There are several directions we will be exploring with this technique. For example, we are interested in using the next frame prediction model to assess the quality of individual frames, particularly in the context of motion and other artifacts. Another possible application could be to use this model to impute frames that have been discarded for motion scrubbing. Finally, we believe unsupervised models can offer novel insights into the dynamics of resting state fluctuations.

\bibliography{unsup}
\bibliographystyle{plain}









\end{document}